\pdfoutput=1

\documentclass[11pt]{article}
\usepackage{graphicx}
\usepackage{caption}   
\usepackage{booktabs} 
\usepackage{multirow} 
\usepackage{tabularx} 
\usepackage{makecell} 
\usepackage{caption}  
\usepackage{amsmath}

\usepackage{amssymb}
\usepackage{mathtools}

\usepackage{ragged2e}
\usepackage[table]{xcolor}
\usepackage{adjustbox}
\usepackage{array}

\definecolor{groupcolor}{RGB}{200,200,200}
\usepackage[most]{tcolorbox}

\tcbset{
  example/.style={
    enhanced,
    colback=white!95!black,
    colframe=blue!70!black,
    fonttitle=\bfseries,
    title=#1,
    attach boxed title to top left={yshift=-2mm},
    boxed title style={colframe=blue!70!black},
    sharp corners,
    boxrule=0.8pt,
    left=5pt,right=5pt,top=5pt,bottom=5pt
  }
}

\usepackage[preprint]{acl}

\usepackage{times}
\usepackage{latexsym}

\usepackage[T1]{fontenc}

\usepackage[utf8]{inputenc}

\usepackage{microtype}

\usepackage{inconsolata}

\usepackage{graphicx}

\title{Self-Route: Automatic Mode Switching via Capability Estimation for Efficient Reasoning}

\author{
  {\bfseries Yang He, Xiao Ding, Bibo Cai, Yufei Zhang, Kai Xiong} \\
  {\bfseries Zhouhao Sun, Bing Qin, Ting Liu} \\
  Research Center for Social Computing and Interactive Robotics \\
  Harbin Institute of Technology, China \\
  \{yhe, xding, bbcai, yfzhang, kxiong, zhsun, qinb, tliu\}@ir.hit.edu.cn
}

\begin{document}
\maketitle
\begin{abstract}
While reasoning-augmented large language models (RLLMs) significantly enhance complex task performance through extended reasoning chains, they inevitably introduce substantial unnecessary token consumption—particularly for simpler problems where Short Chain-of-Thought (Short CoT) suffices. This ``overthinking'' phenomenon leads to inefficient resource usage without proportional accuracy gains. To address this issue, we propose \textit{Self-Route}, a dynamic reasoning framework that automatically selects between general and reasoning modes based on model capability estimation. Our approach introduces a lightweight pre-inference stage to extract capability-aware embeddings from hidden layer representations, enabling real-time evaluation of the model’s ability to solve problems. We further construct \textit{Gradient-10K}, a model difficulty estimation-based dataset with dense complexity sampling, to train the router for precise capability boundary detection. Extensive experiments demonstrate that Self-Route achieves comparable accuracy to reasoning models while reducing token consumption by 30-55\% across diverse benchmarks. The proposed framework demonstrates consistent effectiveness across models with different parameter scales and reasoning paradigms, highlighting its general applicability and practical value.

\end{abstract}

\section{Introduction}

\begin{figure}[t]
  \includegraphics[width=\columnwidth]{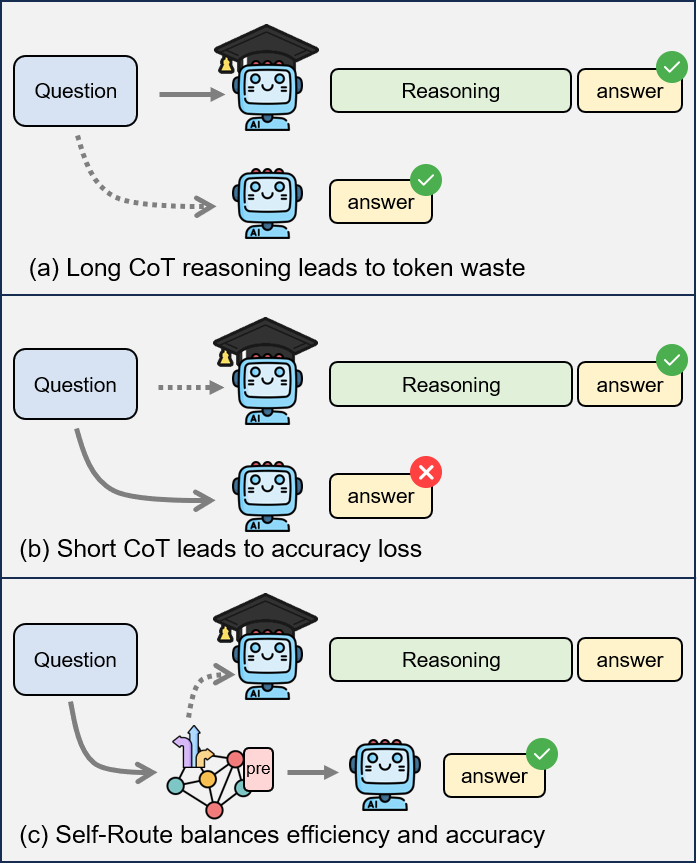}
  \caption{Illustration of the problem scenario addressed by our method, highlighting the trade-offs between Long CoT Reasoning, Short CoT, and the Self-Route approach in balancing efficiency and accuracy.}
  \label{fig:analysis}
\end{figure}

Large language models (LLMs) have demonstrated remarkable capabilities across a wide range of natural language tasks. They have achieved excellent performance in open-ended conversations, text summarization, code generation, and mathematical reasoning~\cite{achiam2023gpt, bubeck2023sparks}. Recently, a new category of models—reasoning large language models (RLLMs), such as OpenAI's O1~\cite{jaech2024openai} and DeepSeek's R1~\cite{guo2025deepseek}, has emerged and significantly improved performance on complex tasks like mathematical reasoning, programming, and interdisciplinary knowledge reasoning~\cite{team2025kimi, chen2025evaluating}. These models leverage  Long Chain-of-Thought (Long CoT) through test-time scaling techniques, generating more detailed intermediate steps during inference, including multi-step thinking, iterative exploration, and reflection~\cite{chen2025towards}. This approach greatly enhances the System-2 reasoning capabilities of LLMs~\cite{li2502system}.

However, Long CoT also comes with certain drawbacks. While they improve accuracy on complex tasks, the multi-step iterative nature of such reasoning often results in extensive outputs, increasing computational costs and inference time. More importantly, for simple problems, Short CoT are usually sufficient to achieve correct and efficient solutions. In such cases, using Long CoT reasoning may lead to ``overthinking''~\cite{chen2024not}, a phenomenon where the model generates thousands of tokens unnecessarily, performing overly complex reasoning that is not proportionate to the problem's difficulty, as shown in ~\autoref{fig:analysis}.

To address the issue of excessive and unnecessary token consumption caused by reasoning models, as well as the limited reasoning capabilities of general models, we propose the \textit{Self-Route} framework. The core objective of this framework is to utilize the Router to select the most efficient reasoning strategy while ensuring correctness.
Our approach introduces a lightweight \textit{pre-inference} stage prior to formal reasoning. By collecting hidden layer representations of the general model during this phase, we construct a capability-aware embedding that reflects the model's capacity to solve the given question. This representation serves as the foundation for the subsequent router.

In order for the router to accurately perceive the boundary of a model’s reasoning capability during inference, 
we further construct a model difficulty estimation-based gradient dataset named \textit{Gradient-10K} to train the router, which contains densely sampled question instances across a continuous spectrum of complexity.
The rationale behind this approach stems from our experimental observation that datasets lacking a difficulty gradient lead to the router's inability to accurately detect the model's capability boundaries, resulting in a significant drop in accuracy.
Training on \textit{Gradient-10K} enables the routing module to make reliable decisions by precisely capturing the interplay between question difficulty and model competence.



We conduct extensive experiments across models of varying scales and architectures to evaluate the effectiveness of our Self-Route framework. The results demonstrate that our method \textbf{significantly reduces token consumption by an average of 30\% to 55\%} across all evaluated datasets compared to reasoning models. Meanwhile, Self-Route maintains a strong accuracy level, with \textbf{less than a 2\% accuracy loss} compared to reasoning models, while achieving significantly higher accuracy than general models. Overall, our method achieves satisfactory performance in estimating model reasoning capability while greatly improving inference efficiency.


Moreover, our approach is not limited to routing across distinct models, it also works well within hybrid reasoning models where Short- and Long-CoT capabilities coexist. This versatility underscores its broad applicability and practical potential in real-world settings.

\begin{figure*}[ht]
    \centering
    \includegraphics[width=\textwidth]{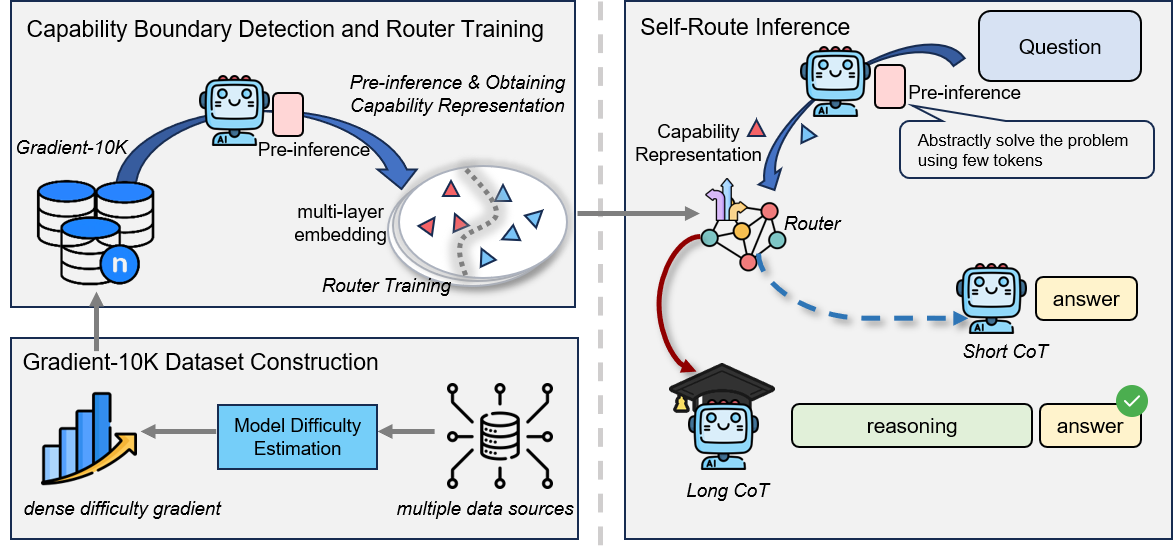} 
    \caption{Workflow from Gradient-10K dataset construction to Self-Route Inference. During capability boundary detection, the pre-inference module collects the hidden layer vector of the last token as the capability representation. The dense difficulty gradient of Gradient-10K ensures accurate reasoning mode selection, balancing correctness and efficiency in Self-Route Inference.}
    \label{fig:self-route}
\end{figure*}

\section{Methods}

\subsection{Overview}

Existing approaches rely on either manually selecting a Short CoT or Long CoT strategy for question answering, as shown in~\autoref{fig:analysis}. 
Short CoT is efficient but struggles with complex questions, while Long CoT reasoning is more accurate but consumes more tokens.
In this work, we propose our Self-Route framework, which introduces a routing module to automatically select the most efficient reasoning strategy while ensuring correctness. 

Specifically, the Self-Route framework consists of three components: the \textbf{Router} \( R \), the \textbf{general model} \( \mathcal{M}_s \), and the \textbf{reasoning model} \( \mathcal{M}_\ell \). For a given query \( q \), the Self-Route framework first performs a \textbf{pre-inference step}: the model reasons about the query \( q \) to extract intermediate hidden representations from various layers. Based on these representations, the trained router estimates the probability that the general model (i.e., \( \mathcal{M}_s \)) will generate a correct answer for \( q \), denoted as \( P(q) \). The routing function then dynamically selects strategies as follows: when \( P(q) \geq \tau \) (indicating the model predicts the Short CoT strategy can solve the query correctly), \( \mathcal{M}_s \) is invoked; otherwise, the system falls back to the Long CoT reasoning strategy \( \mathcal{M}_\ell \), thereby ensuring the correctness of the output.

To effectively train a router that could accurately detect the capability boundaries of different models, we carefully constructed a densely graded difficulty dataset, \textit{Gradient-10K}, based on model difficulty estimation from multiple data sources. Subsequent experiments demonstrate that the dense difficulty gradient plays a crucial role in maintaining high accuracy in capability detection within the Self-Route framework. Training the router on data without the difficulty gradient characteristic leads to a significant drop in inference accuracy due to incorrect capability boundary detection, demonstrating the effectiveness of the constructed Gradient-10K dataset with its dense difficulty gradient. The following sections provide a detailed introduction to each component of the method.


\subsection{Dense Difficulty Gradient Dataset Construction}

To enable the model to fully understand the boundaries of its own capabilities with internal state~\cite{chen2024teaching, li2024knowledge}, specifically to distinguish between problems it can solve within its ability range and those that require deep reasoning through multiple intermediate steps, it is essential to test the model using a dataset with a well-structured difficulty gradient.

A difficulty-gradient dataset refers to a collection of problems that span various difficulty levels, from the simplest to the most challenging. 
The denser and more fine-grained the gradient is, the more effectively the model can explore and understand the limits of its capabilities. In contrast, data lacking a difficulty gradient can lead to issues such as incorrect detection of the model’s capability boundaries, causing a significant drop in inference accuracy.

To construct such a dataset, we collected questions of varying difficulty from multiple sources. To quantify the difficulty level of each question, we employed the accuracy of a general Model on that question as an empirical proxy. Specifically, we define the following scoring function:

\begin{equation}
\mathcal{D}_m(q) = 1 - \mathcal{A}_m(q)
\end{equation}

where $ \mathcal{A}_m(q) \in [0, 1] $ denotes the accuracy of model $ m $ on question $ q $, and $ \mathcal{D}_m(q) \in [0, 1] $ represents the corresponding difficulty score for that model-question pair. A higher value indicates a more challenging question with respect to model $ m $'s reasoning capability.
This formulation allows us to categorize questions into different difficulty levels in a principled manner, facilitating balanced dataset construction across sources and complexities.
We divided the difficulty range into five levels, from easiest to hardest, and constructed the final dataset, \textit{Gradient-10K}, as presented in \autoref{tab:gradient10k-accuracy}.



\begin{table}[t]
\centering
\begin{tabular}{lccc}
\hline
\textbf{Level} & \textbf{$\mathcal{D}_{\text{Q-7B}}(q)$} & \textbf{$\mathcal{D}_{\text{Q-32B}}(q)$} & \textbf{$\mathcal{D}_{\text{Q3-8B}}(q)$} \\
\hline
1  & 0.06 & 0.04 & 0.05 \\
2 & 0.20 & 0.14 & 0.15 \\
3 & 0.40 & 0.30 & 0.27 \\
4 & 0.58 & 0.46 & 0.40 \\
5  & 0.90 & 0.86 & 0.79 \\
\hline
\end{tabular}%
\caption{Estimated average difficulty scores $\mathcal{D}_m(q)$ in the \textit{Gradient-10K} dataset across multiple models. Each score is computed as $ \mathcal{D}_m(q) = 1 - \mathcal{A}_m(q) $, where $ \mathcal{A}_m(q) $ denotes the accuracy of model $ m $ on question $ q $. Q-7B and Q-32B refer to models from the Qwen2.5 series, while Q3-8B corresponds to Qwen3-8B with \texttt{no\_think} mode.}
\label{tab:gradient10k-accuracy}
\end{table}

\subsection{Capability Representation Acquisition through Pre-inference}

Language models can estimate both the difficulty of a question and their own ability to answer it during the generation process~\cite{ashok2025language}. This pre-inference stage enables the model to perform such an assessment using only a small number of tokens, allowing for an efficient capability estimation before committing to a full reasoning path.

For clarity, we provide an example to illustrate the question and pre-inference process as follows:

\begin{tcolorbox}[example=Example]
Let $x, y$ and $z$ be positive real numbers that satisfy:
$$
\log_2\left(\frac{x}{yz}\right) = \frac{1}{2},  \log_2\left(\frac{y}{xz}\right) = \frac{1}{3},
$$
$$
\log_2\left(\frac{z}{xy}\right) = \frac{1}{4}.
$$
Find $\left|\log_2(x^4y^3z^2)\right| = \frac{m}{n}$ where $m$ and $n$ are coprime positive integers. Compute $m+n$.
\\
\\
\textbf{\textit{Pre-inference:}} Convert logarithmic equations to exponential form to express $x, y,$ and $z$ in terms of each other. Solve the system of equations to find ratios of $x, y, z$. Calculate $\log_2(x^4y^3z^2)$, simplify the expression, take absolute value. Express result as $\frac{m}{n}$ in simplest form and find $m+n$.
\end{tcolorbox}

Through the \textit{Pre-inference} process, the model attempts to solve the problem from a global perspective, adopting an abstract and generalized reasoning approach. This process essentially aligns the problem's actual difficulty with the model's internal problem-solving capability, with the hidden representations serving as a compact and informative capability-aware embedding.

Formally, given a question $ q $, we first generate a Short CoT trajectory $ r $ under a limited token budget $ \tau $, using the general model $ m = \mathcal{M}_s $. The generation is defined as:

\begin{equation}
r = G_{\mathcal{M}_s}(q; \tau)
\end{equation}

$ G_{\mathcal{M}_s}(\cdot; \tau) $ denotes the generation function of the general model; $ \tau $ is the maximum number of tokens allowed in the pre-inference step.

Based on this generated response $ r $, we collect the hidden representations from all intermediate transformer layers at the final decoding timestep $ T $. Let $ L $ denote the number of layers. Under the condition that $ r $ has been generated from $ q $, the capability-aware representation is defined as:

\begin{equation}
\mathbf{H}_\text{cap}(q) = \left\{ \mathbf{h}^{(1)}_T(q \mid r),\ \dots,\ \mathbf{h}^{(L)}_T(q \mid r) \right\}
\end{equation}


where $ \mathbf{h}^{(l)}_T(q \mid r) \in \mathbb{R}^d $ denotes the hidden state at layer $ l $, conditioned on the input question $ q $ and derived under the context of the generated response $ r $, at the final timestep $ T $. This representation captures the model's internal assessment of its own capability to address the problem.

\subsection{Router Training}

The role of the router is to decide whether to use a reasoning model or a general model for the current problem. The key is to enable the model to fully assess whether it can solve the problem within its Short CoT reasoning capability. Using linear functions can effectively determine the model's future output behavior~\cite{ashok2025language}. The task of exploring the model’s capability boundaries is made possible through the pre-inference vector representations collected on the Gradient-10K difficulty gradient dataset. 

We utilize the hidden layer vector representation of the final output token from the \textit{Pre-inference} module to train the router, which estimates the model's ability to solve a given problem. The router uses a linear function as follows:

\begin{equation}
\hat{y}_i = \mathbf{w}^\top \mathbf{h}^{(l)}_T + b
\end{equation}

where
 \( \mathbf{h}^{(l)}_T \) is the hidden vector of the last token at layer \( l \),
 \( \mathbf{w} \) is the weight vector,
 \( b \) is the bias term,
 \( \hat{y}_i \) is the predicted label for problem \( i \).

The model is trained by minimizing the cross-entropy loss:


\begin{equation}
\scalebox{0.85}{$\displaystyle
\mathcal{L} = - \sum_{i=1}^{N} \left[ y_i \log(\hat{y}_i) + (1 - y_i) \log(1 - \hat{y}_i) \right]
$}
\end{equation}

where \( y_i \) is the true solution correctness label, and \( \hat{y}_i \) is the predicted label.

\section{Experiments}

\begin{table*}[t]
\centering
\footnotesize
\renewcommand{\arraystretch}{1.15}
\setlength{\defaultaddspace}{0.25em}
\adjustbox{max width=\textwidth}{%
\begin{tabularx}{1.20\textwidth}{l *{6}{rr}} 
\toprule[1.2pt]
\rowcolor{white}
\multirow{2}{*}{\textbf{Model}} & 
\multicolumn{2}{c}{\textbf{GSM8K}} & 
\multicolumn{2}{c}{\textbf{Math500}} & 
\multicolumn{2}{c}{\textbf{\makecell{GPQA\\Diamond}}} & 
\multicolumn{2}{c}{\textbf{\makecell{AIME-2024}}} & 
\multicolumn{2}{c}{\textbf{\makecell{ARC\\Challenge}}} & 
\multicolumn{2}{c}{\textbf{AVG}} \\
\cmidrule(lr){2-3} \cmidrule(lr){4-5} \cmidrule(lr){6-7} \cmidrule(lr){8-9} \cmidrule(lr){10-11} \cmidrule(lr){12-13}
& \textbf{Acc} & \textbf{Token} & \textbf{Acc} & \textbf{Token} & \textbf{Acc} & \textbf{Token} & \textbf{Acc} & \textbf{Token} & \textbf{Acc} & \textbf{Token} & \textbf{Acc} & \textbf{Token} \\
\midrule[1pt]

\rowcolor{groupcolor}
\multicolumn{13}{c}{\textbf{Reasoning Model Comparison}} \\
Marco-O1 & 88.9 & 622.1 & 72.0 & 1405.5 & 26.3 & 1580.8 & 10.0 & 7229.1 & 88.1 & 516.5 & 57.1 & 825.9 \\
OpenO1-Qwen-7B & 90.0 & 575.2 & 71.8 & 1157.4 & 27.3 & 1149.6 & 6.7 & 2046.7 & 88.4 & 513.0 & 56.8 & 692.0 \\
Skywork-O1 & 91.1 & 323.2 & 78.4 & 850.5 & 23.7 & 1682.3 & 16.7 & 2607.5 & 71.3 & 452.7 & 56.2 & 557.1 \\
Bespoke-Stratos-7B & 93.1 & 1075.0 & 85.4 & 4169.1 & 38.9 & 4405.3 & 13.3 & 13176.7 & 91.3 & 928.0 & 64.4 & 1819.7 \\
\rowcolor{groupcolor}
\multicolumn{13}{c}{\textbf{Self-Route on Qwen2.5-7B}} \\
\multicolumn{13}{l}{\textbf{\textit{Genaral Model}}} \\
Qwen2.5-7B-Instruct & 92.3 & 308.4 & 77.8 & 626.4 & 31.8 & 1470.6 & 13.3 & 1219.9 & 90.9 & 361.2 & 61.2 & 457.0 \\
\multicolumn{13}{l}{\textbf{\textit{Reasoning Model}}} \\
R1-Distill-Qwen-7B & 91.2 & 1225.8 & 95.0 & 3922.9 & 40.4 & 17654.0 & 56.7 & 20021.4 & 83.1 & 1328.8 & 73.3 & 2867.8 \\
OpenR1-Qwen-7B & 96.1 & 1282.6 & 93.4 & 3704.1 & 41.4 & 8759.7 & 53.3 & 11664.1 & 80.9 & 1862.7 & 73.0 & 2426.6 \\
Gradient-7B & 91.0 & 1196.9 & 83.2 & 5410.9 & 35.9 & 16004.8 & 56.3 & 17683.6 & 90.5 & 1833.5 & 62.8 & 2815.6 \\
\multicolumn{13}{l}{\textbf{\textit{Self-Route Method}}} \\
R1-Distill-Qwen-7B\textsubscript{Self-Route} & 92.3 & 324.3 & 87.8 & 2513.8 & 37.4 & 17225.5 & 56.7 & 19622.2 & 89.2 & 648.2 & 72.7 & \textbf{2001.8 (↓30\%)} \\
OpenR1-Qwen-7B\textsubscript{Self-Route} & 92.6 & 319.3 & 87.2 & 2294.3 & 39.9 & 3207.4 & 53.3 & 11356.8 & 88.7 & 691.6 & 72.3 & \textbf{1353.8 (↓44\%)} \\
Gradient-7B\textsubscript{Self-Route} & 92.3 & 326.8 & 80.8 & 3360.8 & 36.5 & 9642.1 & 56.3 & 17377.1 & 90.4 & 726.8 & 62.3 & \textbf{1669.5 (↓41\%)} \\
\rowcolor{groupcolor}
\multicolumn{13}{c}{\textbf{Self-Route on Qwen2.5-32B}} \\
\multicolumn{13}{l}{\textbf{\textit{Genaral Model}}} \\
Qwen2.5-32B-Instruct & 96.3 & 312.4 & 84.4 & 583.2 & 50.5 & 1080.4 & 23.3 & 1020.8 & 95.3 & 337.5 & 70.0 & 417.4 \\
\multicolumn{13}{l}{\textbf{\textit{Reasoning Model}}} \\
R1-Distill-Qwen-32B & 95.7 & 637.3 & 96.0 & 3344.3 & 63.1 & 6490.9 & 73.3 & 8770.6 & 96.1 & 599.3 & 84.8 & 1478.2 \\
\multicolumn{13}{l}{\textbf{\textit{Self-Route Method}}} \\
R1-Distill-Qwen-32B\textsubscript{Self-Route} & 96.4 & 316.6 & 93.6 & 1859.8 & 61.1 & 6242.4 & 73.3 & 8523.6 & 95.6 & 376.0 & 84.0 & \textbf{1018.9 (↓31\%)}\\
\rowcolor{groupcolor}
\multicolumn{13}{c}{\textbf{Self-Route on Qwen3-8B}} \\
\multicolumn{13}{l}{\textbf{\textit{Genaral Model}}} \\
Qwen3-8B\textsubscript{no\_think} & 94.0 & 278.2 & 85.6 & 1072.9 & 49.5 & 1775.7 & 23.3 & 5058.9 & 94.4 & 394.5 & 69.4 & 580.7 \\
\multicolumn{13}{l}{\textbf{\textit{Reasoning Model}}} \\
Qwen3-8B\textsubscript{think} & 96.5 & 2049.5 & 97.0 & 5152.3 & 56.6 & 9473.6 & 76.7 & 14848.5 & 95.7 & 1346.4 & 84.5 & 2851.4 \\
\multicolumn{13}{l}{\textbf{\textit{Self-Route Method}}} \\
Qwen3-8B\textsubscript{Self-Route} & 94.3 & 332.8 & 91.4 & 2376.4 & 53.6 & 8430.8 & 76.7 & 13857.7 & 94.4 & 427.8 & 82.6 & \textbf{1309.0 (↓54\%)} \\
\bottomrule[1.2pt]
\end{tabularx}}
\vspace{-3pt}
\caption{Accuracy and token consumption performance of reasoning models, general models, and Self-Route across various problem types and difficulty levels, along with a comparison of reasoning model effects.}
\label{tab:model-compare}
\end{table*}

\subsection{Experimental Setup}

We evaluate the effectiveness of the Self-Route method on models with various parameter configurations and inference modes. The general models include Qwen2.5-7B-Instruct, Qwen2.5-32B-Instruct~\cite{yang2024qwen2}, and the latest Qwen3-8B model. For Long CoT reasoning, we select R1-Distill-Qwen-7B, OpenR1-Qwen-7B~\cite{guo2025deepseek}, and Gradient-7B as the reasoning models for Qwen2.5-7B-Instruct routing; R1-Distill-Qwen-32B as the reasoning model for Qwen2.5-32B-Instruct routing; and Qwen3-8B (think mode) as the reasoning model for Qwen3-8B (no\_think mode) routing. This experimental setup thoroughly explores routing strategies across models with different parameter scales and reasoning mechanisms, as well as the feasibility and robustness of the Self-Route method when using hybrid reasoning models (Qwen3-8B). Gradient-7B is based on Qwen2.5-7B-Instruct and trained on the Gradient-10K dataset by distilling reasoning trajectories from DeepSeek-R1.

\textbf{Training Dataset.} In this experiment, all model training—including all size of routers, was conducted on our constructed Gradient-10K dataset. Based on the principle of difficulty grading across multiple data sources, we integrated a variety of datasets with different levels of complexity, including Orca Math~\cite{mitra2024orca}, AMC AIME, Olympiads, GSM8K~\cite{cobbe2021training}, and s1k~\cite{muennighoff2025s1}. 

\textbf{Evaluation Dataset.} For the test datasets, we selected a variety of commonly used mathematical test datasets with different difficulty levels, domain-specific knowledge QA datasets, and science question-answering datasets. (1) \textbf{GPQA}~\cite{rein2024gpqa} is a PhD-level science multiple-choice QA dataset. The questions are authored by domain experts in physics, chemistry, and biology. In our main experiments, we use the highest quality diamond set containing 198 questions. (2) \textbf{Math benchmarks}  include \textbf{GSM8K}~\cite{cobbe2021training}, \textbf{MATH500}~\cite{lightman2023let}, \textbf{AIME-2024}. MATH500 consists of 500 questions from the MATH test set. AIME-2024 are middle school math competitions covering arithmetic, algebra, geometry, etc., containing 30 questions. (3) \textbf{ARC-Challenge}~\cite{clark2018think} is a subset of the AI2-ARC dataset, featuring difficult science questions designed to challenge advanced question-answering systems. In this paper, we test the ARC dataset by treating the multiple-choice options as part of the question, following the multiple-choice question format~\cite{borchmann2024case}.

\subsection{Main Results}

The results in \autoref{tab:model-compare} provide a comprehensive comparison between our Self-Route method and various reasoning and baseline models. By analyzing datasets of different difficulty levels, we observe a consistent pattern across models with varying parameter scales and architectures: 

(i) For relatively simple tasks where the general model already performs well, such as on GSM8K or ARC-Challenge, the routing mechanism tends to favor Short CoT reasoning paths. As a result, Self-Route solves the majority of such questions by routing them to the general model for reasoning, thereby achieving strong answer performance while significantly reducing token consumption compared to Long CoT reasoning approaches.

(ii) In contrast, for more complex tasks where the general model struggles, such as GPQA Diamond or AIME-2024, Self-Route predominantly selects the Long CoT reasoning path. This leads to a notable improvement in accuracy compared to the general model, despite the increased token consumption associated with extended reasoning processes.

To provide an overall assessment across all datasets, we compute the average accuracy and a weighted average of token consumption based on dataset size. As shown in the \autoref{tab:model-compare}, Self-Route consistently achieves performance on par with various Long CoT reasoning methods, suffering \textbf{less than 1\% drop in average accuracy} on Qwen2.5-7B and Qwen2.5-32B, while \textbf{reducing token consumption by 30\% to 55\%} across Qwen2.5-7B, Qwen2.5-32B, and Qwen3-8B. These results demonstrate that Self-Route can achieve substantial efficiency gains at only a marginal cost to accuracy.

Furthermore, the applicability of Self-Route extends beyond cross-model routing (e.g., between general and reasoning models). It also proves effective within a single hybrid reason model, such as Qwen3, for dynamically switching between internal reasoning modes. This adaptability highlights the potential of Self-Route for future deployment in unified model architectures, enabling intelligent and automated inference routing based on the problem complexity.

\subsection{Exploring the Necessity of a Dense Difficulty Gradient}

In this section, we investigate the necessity of constructing a densely annotated difficulty gradient dataset. We compare two approaches for characterizing model capabilities and training routers: one using our constructed Model Difficulty Estimation-Based Gradient Dataset—Gradient-10K—and the other relying on the GSM8K training set without an explicit difficulty gradient. The overall performance is summarized in ~\autoref{tab:router-model-compare}. As shown, the router trained on Gradient-10K achieves significantly better accuracy across the majority of test sets compared to its GSM8K-based counterpart. The router trained on GSM8K may \textbf{cause up to an approximately 11\% decrease in accuracy}. This indicates that the lack of a dense difficulty gradient leads to inaccurate estimation of the model's capability boundary, which in turn results in a substantial drop in routing accuracy.

\begin{table}[t]
\centering
\renewcommand{\arraystretch}{1.2}
\resizebox{\columnwidth}{!}{%
\begin{tabular}{lccc}
\hline
\textbf{Model} & \textbf{Acc.} & \textbf{Prec.} & \textbf{F1} \\
\hline
\multicolumn{4}{c}{\textbf{\textit{Router on Gradient}}} \\
\hline
Qwen2.5-7B\textsubscript{Gradient-Router} & 0.81 & 0.83 & \textbf{0.86} \\
Qwen2.5-32B\textsubscript{Gradient-Router} & 0.84 & 0.89 & \textbf{0.89} \\
Qwen3-8B\textsubscript{Gradient-Router} & 0.83 & 0.88 & \textbf{0.89} \\
\hline
\multicolumn{4}{c}{\textbf{\textit{Router on GSM8K}}} \\
\hline
Qwen2.5-7B\textsubscript{GSM8K-Router} & 0.70 & 0.70 & 0.80 \\
Qwen2.5-32B\textsubscript{GSM8K-Router} & 0.77 & 0.81 & 0.85 \\
Qwen3-8B\textsubscript{GSM8K-Router} & 0.78 & 0.79 & 0.87 \\
\hline
\end{tabular}%
}
\caption{Comparison of evaluation metrics on the Gradient Test set between the router trained with a dense difficulty gradient and the router trained on the gradient-free GSM8K dataset.}
\label{tab:router-metrics-compare}
\end{table}

\begin{table*}[ht]
\centering
\footnotesize
\renewcommand{\arraystretch}{1.15}
\setlength{\defaultaddspace}{0.25em}
\adjustbox{max width=\textwidth}{%
\begin{tabularx}{1.15\textwidth}{l *{6}{rr}} 
\toprule[1.2pt]
\rowcolor{white}
\multirow{2}{*}{\textbf{Model}} & 
\multicolumn{2}{c}{\textbf{GSM8K}} & 
\multicolumn{2}{c}{\textbf{Math500}} & 
\multicolumn{2}{c}{\textbf{\makecell{GPQA\\Diamond}}} & 
\multicolumn{2}{c}{\textbf{\makecell{AIME\\2024}}} & 
\multicolumn{2}{c}{\textbf{\makecell{ARC\\Challenge}}} & 
\multicolumn{2}{c}{\textbf{AVG}} \\
\cmidrule(lr){2-3} \cmidrule(lr){4-5} \cmidrule(lr){6-7} \cmidrule(lr){8-9} \cmidrule(lr){10-11} \cmidrule(lr){12-13}
& \textbf{Acc} & \textbf{Token} & \textbf{Acc} & \textbf{Token} & \textbf{Acc} & \textbf{Token} & \textbf{Acc} & \textbf{Token} & \textbf{Acc} & \textbf{Token} & \textbf{Acc} & \textbf{Token} \\
\midrule[1pt]
\multicolumn{13}{c}{\textbf{\textit{Router on Gradient}}} \\
\hline
R1-Distill-Qwen-7B\textsubscript{Gradient-Router} & \textbf{92.3} & 324.3 & \textbf{87.8} & 2513.8 & \textbf{37.4} & 17225.5 & \textbf{56.7} & 19622.2 & 89.2 & 648.2 & \textbf{72.7} & 2001.8 \\
R1-Distill-Qwen-32B\textsubscript{Gradient-Router} & \textbf{96.4} & 316.6 & \textbf{93.6} & 1859.8 & \textbf{61.1} & 6242.4 & \textbf{73.3} & 8523.6 & \textbf{95.6} & 376.0 & \textbf{84.0} & 1018.9 \\
Qwen3-8B\textsubscript{Gradient-Router} & \textbf{94.3} & 332.8 & \textbf{91.4} & 2376.4 & \textbf{53.6} & 8430.8 & \textbf{76.7} & 13857.7 & 94.4 & 427.8 & \textbf{82.6} & 1309.0 \\
\hline
\multicolumn{13}{c}{\textbf{\textit{Router on GSM8K}}} \\
\hline
R1-Distill-Qwen-7B\textsubscript{GSM8K-Router} & 91.8 & 358.4 & 82.4 & 1746.5 & 36.4 & 8036.5 & 30.0 & 7923.9 & \textbf{90.8} & 362.6 & 66.28 & 1118.3 \\
R1-Distill-Qwen-32B\textsubscript{GSM8K-Router} & 96.3 & 325.9 & 90.8 & 1625.0 & 52.0 & 3227.9 & 53.3 & 5447.9 & 95.2 & 342.3 & 77.52 & 759.9 \\
Qwen3-8B\textsubscript{GSM8K-Router} & 94.1 & 312.2 & 87.4 & 1550.8 & 50.0 & 6816.9 & 33.3 & 6887.3 & \textbf{94.5} & 492.2 & 71.86 & 1031.5 \\
\bottomrule[1.2pt]
\end{tabularx}}
\vspace{-3pt}
\caption{Routing methods trained on the Gradient-10K dataset with a dense difficulty gradient versus routers trained on GSM8K without an explicit difficulty gradient: accuracy and token consumption across various test sets..}
\label{tab:router-model-compare}
\end{table*}

Furthermore, by comparing the two routers on the Gradient test set (as shown in the \autoref{tab:router-metrics-compare}), we observe that the Gradient-Router consistently outperforms the GSM8K-Router in key evaluation metrics. This result corroborates the findings from ~\autoref{tab:router-model-compare}, reinforcing the reliability of our experimental outcomes and highlighting the importance of a dense difficulty gradient in achieving accurate model capability estimation.

\subsection{Pre-inference Analysis}

\subsubsection{Token Consumption in Pre-inference}

In this section, we explore whether the \textit{Pre-inference} module introduces unnecessary additional token consumption. The inference processes tested here include pre-inference, Short CoT and Long CoT reasoning. We quantitatively analyze the additional computational cost introduced by the pre-inference module by comparing its token consumption across datasets of varying difficulty levels and domains, as well as the ratio of pre-inference token usage relative to that of the reasoning model. This analysis helps determine if pre-inference can achieve good performance without causing excessive token consumption for the system.

\begin{table}[t]
\centering
\begin{tabular}{lcccc}
\hline
\textbf{Dataset} & 
\textbf{Short} & 
\textbf{Long} & 
\textbf{Pre} & 
\textbf{Ratio (\%)} \\
\hline
ARC     & 361  & 1329 & 106 &  7.9 \\
GSM8K   & 308  & 1226 &  67 &  5.5 \\
Math500 & 626  & 3923 & 120 &  3.0 \\
GPQA    & 1471 & 17654 & 172 & 1.0 \\
AIME    & 1220 & 20021 & 136 & 0.7 \\
\hline
\end{tabular}
\caption{Token consumption analysis of the pre-inference module on Qwen2.5-7B across reasoning paradigms. "Short" and "Long" refer to the number of tokens used in Short and Long CoT reasoning, respectively. "Pre" indicates the number of tokens consumed by the pre-inference module, and "Ratio" denotes the percentage of Pre over Long.}
\label{tab:pre-infer-assumption}
\end{table}

As shown in \autoref{tab:pre-infer-assumption}, the token consumption of pre-inference is significantly lower compared to that of formal reasoning. With most cases, \textbf{the pre-inference token consumption is less than 5\% of that of Long CoT reasoning}. This demonstrates that pre-inference is a highly cost-effective strategy, enabling efficient routing decisions with minimal resource overhead.

\subsubsection{Analysis of Routing Effectiveness}

This section investigates whether intermediate hidden layer vectors from the Pre-inference module can effectively represent a model’s problem-solving capability. We evaluate which layer provides the best features for routing by training routers on all hidden layers of Qwen2.5-7B and Qwen2.5-32B.

\begin{figure}[t]
\centering
\includegraphics[width=\linewidth]{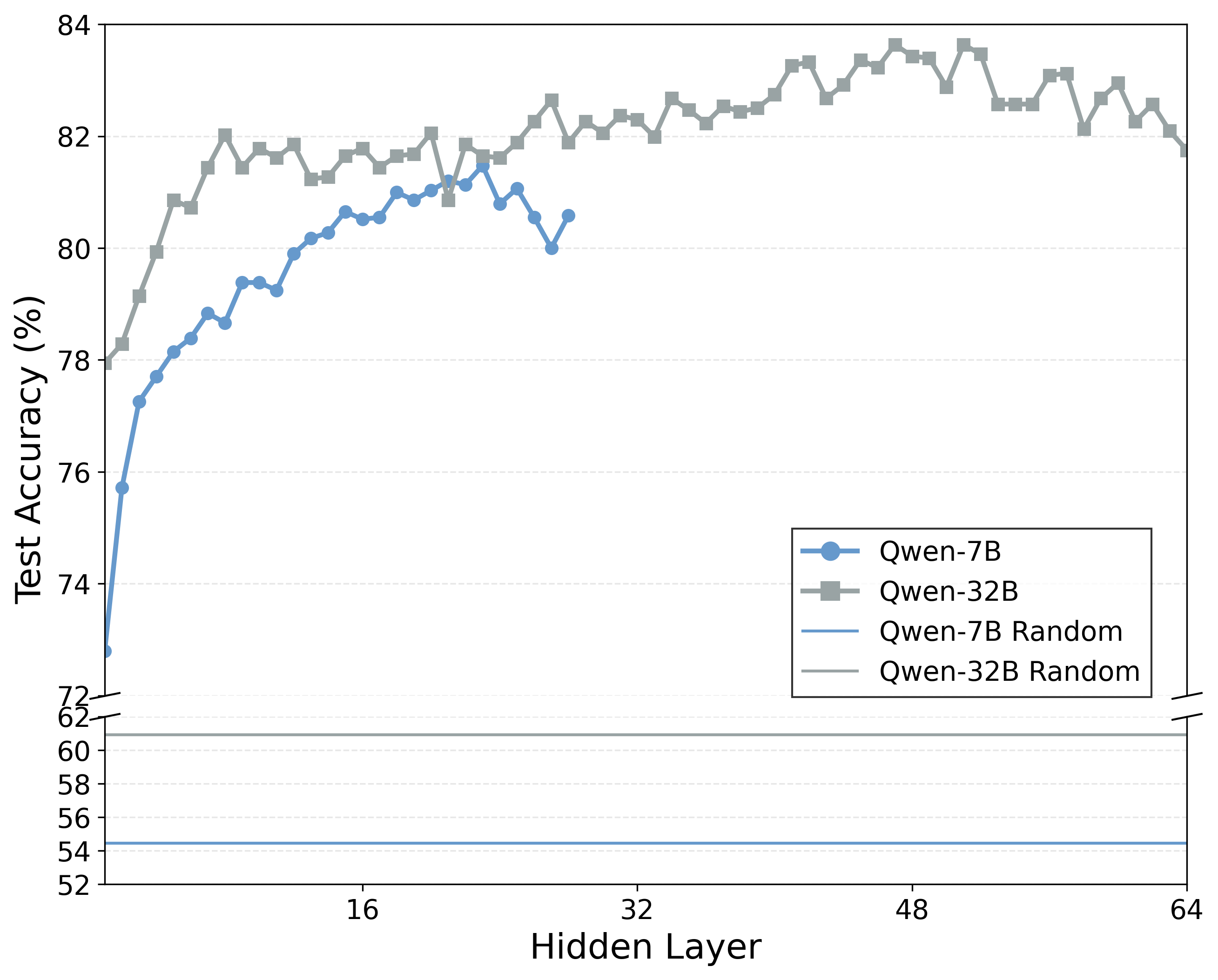}
\caption{Routing accuracy of pre-inference vector representations from different hidden layers for Qwen2.5-7B and Qwen2.5-32B.}
\label{fig:preinfer-layer-analysis}
\end{figure}

From the results depicted in \autoref{fig:preinfer-layer-analysis}, we observe two key patterns:

(i) Utilizing vector representations from appropriate intermediate hidden layers as indicators of the model's problem-solving capability achieves strong performance, enabling an accurate assessment of whether the general model can effectively address the given problem.

(ii) Lower layers tend to capture more concrete semantic features, while higher layers encode increasingly abstract semantic information, such as the model’s ability to solve the problem. Our experiments show that hidden layers around the 60\% – 80\% position in the network typically yield the most effective estimation of model capability.

\subsection{Analysis of the Impact of Self-Route}
This section compares the performance of Qwen2.5, R1-Distill-Qwen-7B, and the proposed Self-Route method in terms of accuracy across datasets of varying difficulty levels.

\begin{figure}[t]
\centering
\includegraphics[width=\linewidth]{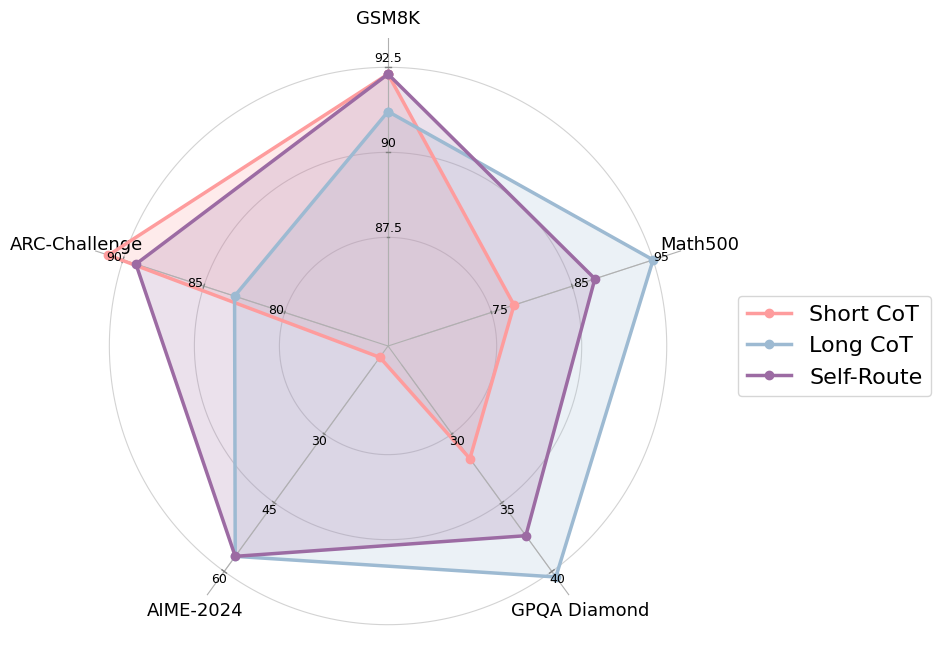}
\caption{Radar chart comparing the accuracy of Qwen2.5-7B, R1-Distill-Qwen-7B, and Self-Route across multiple datasets. Self-Route maintains high accuracy while significantly reducing computational cost.}
\label{fig:accuracy-radar}
\end{figure}

In addition to efficiency improvements, the radar chart presented in \autoref{fig:accuracy-radar} offers a comprehensive comparison of the accuracy achieved by three reasoning approaches across a diverse set of datasets. The results clearly demonstrate that the Self-Route method not only significantly enhances the performance of the Short Chain-of-Thought (Short CoT) reasoning approach but also effectively compensates for the limitations of reasoning models. On the ARC-Challenge dataset, the general model outperforms the reasoning model, whereas in complex mathematical reasoning tasks such as those found in AIME or GPQA, the reasoning model achieves superior performance compared to general models. The Self-Route approach dynamically selects the most appropriate reasoning mode for each question type, thereby optimizing overall performance.


Moreover, as illustrated by both \autoref{fig:accuracy-radar} and \autoref{fig:token-comparison}, Self-Route achieves a balance between accuracy and efficiency by significantly reducing token consumption without compromising answer quality. The approach strategically avoids redundant reasoning steps and excess token usage, resulting in substantial computational savings.

\section{Related Work}

 Recent studies on Model-based Efficient Reasoning have primarily focused on fine-tuning large language models (LLMs) to improve their intrinsic ability to reason concisely and efficiently. Existing works leverage traditional RL optimization techniques combined with explicit length-based reward to control the length of CoT reasoning~\cite{arora2025training, yeo2025demystifying, aggarwal2025l1}. In contrast, supervised fine-tuning (SFT) strategies aim to improve reasoning efficiency by constructing variable-length CoT reasoning datasets. This includes techniques such as post-reasoning CoT compression~\cite{yu2024distilling, kang2025c3ot, xia2025tokenskip} and obtaining compressed CoT data during reasoning~\cite{liu2024can, munkhbat2025self, han2024token}.

Works center on estimating knowledge boundaries from internal model signals, aim to teach LLMs to recognize and express their own knowledge limits~\cite{chen2024teaching}. Closely related efforts include more accurate knowledge perception~\cite{liang2024learning}, aligning response quality with confidence through model calibration~\cite{tao2024trust}, identifying the causes of hallucination while balancing factual alignment and instruction following~\cite{lin2024flame}, and leveraging diverse methods for factuality alignment through internal perception mechanisms~\cite{tian2023fine}.

Regarding the trade-off between model performance and reasoning efficiency, recent studies have proposed efficient routers that dynamically select between stronger or weaker LLMs during inference~\cite{ong2024routellm}. However, these methods have not yet succeeded in utilizing the model's own internal signals to estimate task difficulty or its capability boundaries during reasoning.

\section{Conclusion}

We propose \textit{Self-Route}, a pre-inference paradigm that leverages models' internal capability representations to optimize reasoning efficiency while ensuring correctness. By constructing \textit{Gradient-10K}, a Model Difficulty Estimation-Based Gradient Dataset, we enable precise detection of model capability boundaries. Experiments demonstrate that Self-Route reduces token consumption by 30-55\% compared to reasoning models with <2\% accuracy loss, while outperforming general model baselines. Remarkably, the framework works effectively both for routing between distinct models and within unified hybrid reasoning architectures (e.g., Qwen3), demonstrating broad practical applicability. Our approach provides a lightweight, generalizable solution for deploying efficient yet accurate reasoning LLMs.

\section*{Limitations}

While Self-Route demonstrates promising results, some constraints remain and deserve additional study. First, our current routing mechanism relies on linear functions to map hidden representations to capability estimates. While this ensures computational efficiency, more sophisticated non-linear or attention-based routing modules could better capture nuanced relationships between model states and problem-solving competence. The framework currently operates in a binary mode (Short CoT vs. Long CoT). Future work could extend this to \textit{multi-expert routing}, dynamically selecting from specialized models (e.g., math-focused, code-generation, or logical-reasoning experts) to achieve finer-grained optimization. These expansions would better align with real-world scenarios requiring cross-domain problem-solving while further improving efficiency-accuracy trade-offs.

\bibliography{custom}

\appendix

\section{Appendix}
\label{sec:appendix}

\subsection{Model Capability Estimation}
\begin{figure}[ht]
\centering
\includegraphics[width=\linewidth]{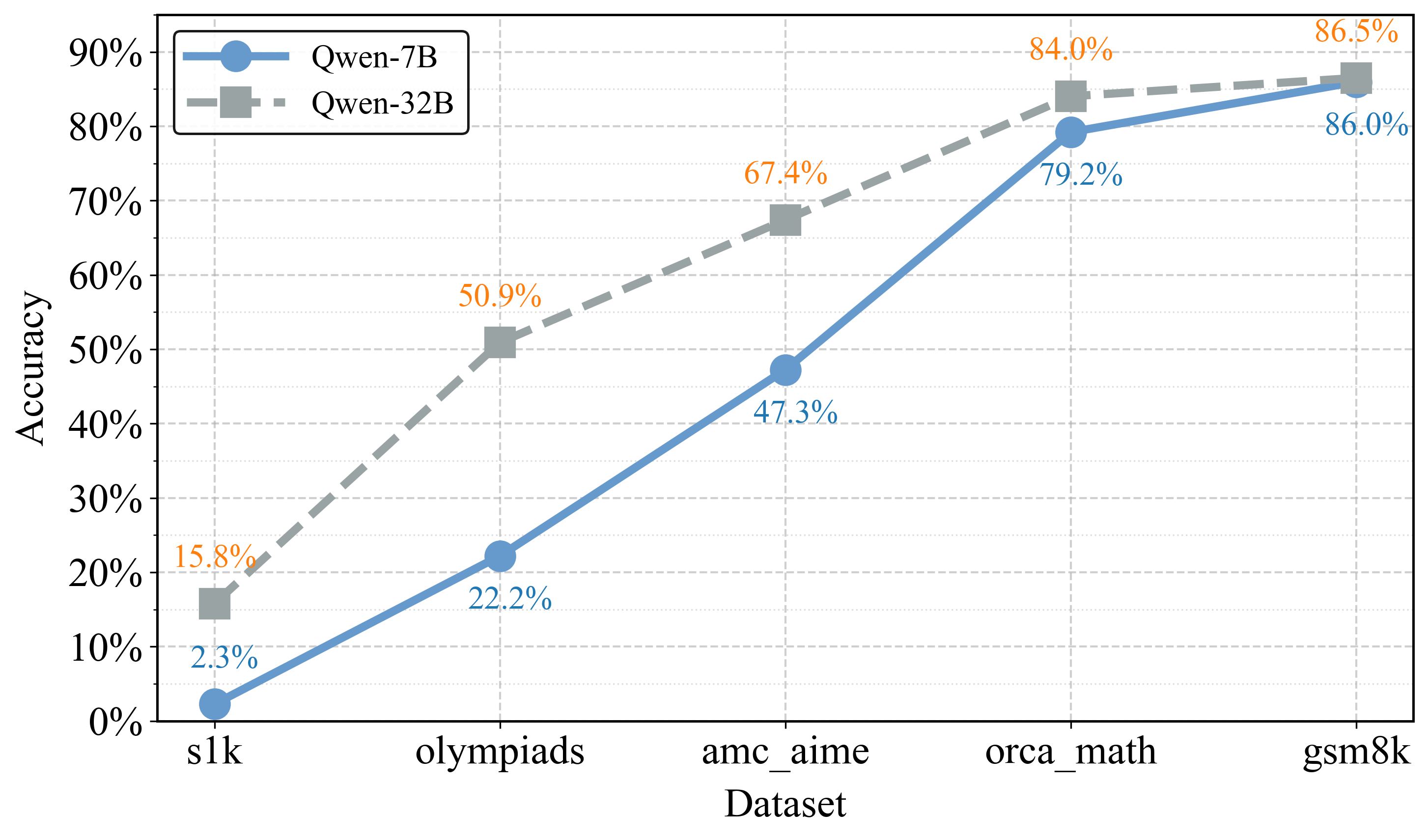}
\caption{Model capability estimation across models at different difficulty levels of the dataset}
\label{fig:preinfer-model-comparison}
\end{figure}

As shown in \autoref{fig:preinfer-model-comparison}, it reveals that models with stronger reasoning capabilities exhibit distinct behavior in estimating their own problem-solving capacity. When applied to data of the same difficulty level, larger models (e.g., Qwen2.5-32B) tend to classify a greater number of questions as within their solvable range, compared to smaller models such as Qwen2.5-7B. This suggests that more powerful models possess a more confident self-assessment mechanism, leading them to make routing decisions that favor Short CoT or direct answering over costly Long CoT reasoning paths.

\subsection{Token Usage Analysis}

\begin{figure}[ht]
\centering
\includegraphics[width=\linewidth]{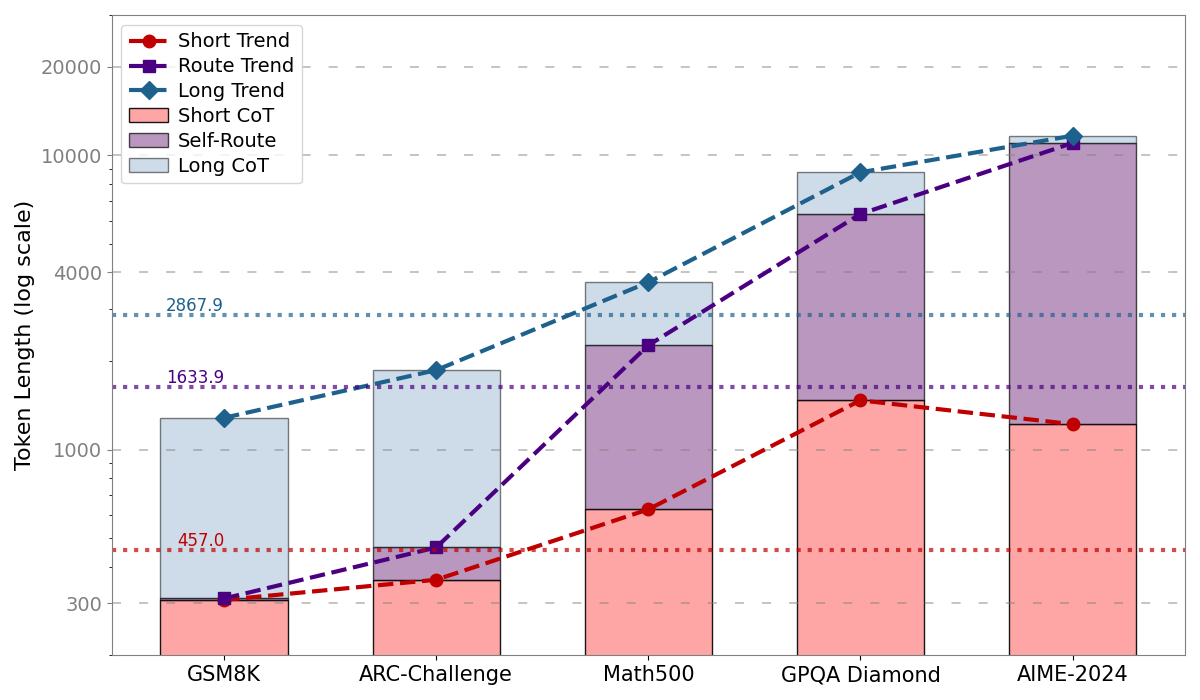}
\caption{Comparison of average token consumption across different reasoning methods (Qwen2.5, R1-Distill-Qwen-7B, and Self-Route) on data with varying difficulty levels. The plot highlights that Self-Route achieves significantly lower token usage compared to reasoning models while maintaining competitive problem-solving capability.}
\label{fig:token-comparison}
\end{figure}

As shown in \autoref{fig:token-comparison}, Self-Route demonstrates a more efficient inference strategy by adaptively choosing between Short and Long CoT reasoning paths based on problem complexity. In contrast, Long CoT reasoning models such as R1-Distill-Qwen-7B consistently consume more tokens across all difficulty levels, even when simpler questions could be resolved with minimal computation.

\subsection{Pre-inference Prompt}

The following prompt is used for pre-inference reasoning path elicitation:

\medskip
\noindent\texttt{Please give a very brief primary plan about how to solve the problem. Just give a very very brief plan, no need for details, calculations or final answer. Just a very brief analysis. Less than 200 words.}
\medskip

\subsection{Router Training Parameters}

The router used in this work is a linear function, with its dimension set to the intermediate hidden layer vector dimension of the model. To identify the most representative hidden layer for model capability characterization, we train the router on all layers of the model. The training parameters for each layer are as follows:

- \textbf{Number of epochs}: 5
    
- \textbf{Batch size}: 32
    
- \textbf{Learning rate}: $1 \times 10^{-4}$

\end{document}